\documentclass[conference]{IEEEtran}
\IEEEoverridecommandlockouts
\usepackage{cite}
\usepackage{amsmath,amssymb,amsfonts}
\usepackage{algorithmic}
\usepackage{graphicx}
\usepackage{textcomp}
\usepackage{booktabs}
\usepackage{diagbox}
\usepackage{xcolor}
\def\BibTeX{{\rm B\kern-.05em{\sc i\kern-.025em b}\kern-.08em
    T\kern-.1667em\lower.7ex\hbox{E}\kern-.125emX}}
\begin{document}

\title{Enhancing Mobile Privacy and Security: A Face Skin Patch-Based Anti-Spoofing Approach
}

\author{Qiushi Guo\\
AI Lab, China Merchants Bank\\
{\tt\small guoqiushi910@cmbchina.com}}

\maketitle

\begin{abstract}
As Facial Recognition System(FRS) is widely applied in areas such as access control and mobile payments due to its convenience and high accuracy. The security of facial recognition is also highly regarded. The Face anti-spoofing system(FAS) for face recognition is an important component used to enhance the security of face recognition systems. Traditional FAS used images containing identity information to detect spoofing traces, however there is a risk of privacy leakage during the transmission and storage of these images. Besides, the encryption and decryption of these privacy-sensitive data takes too long compared to inference time by FAS model. To address the above issues, we propose a face anti-spoofing algorithm based on facial skin patches leveraging pure facial skin patch images as input, which contain no privacy information, no encryption or decryption is needed for these images. We conduct experiments on several public datasets, the results prove that our algorithm has demonstrated superiority in both accuracy and speed.
\end{abstract}

\section{Introduction}

With the advent of deep learning and the widespread use of mobile devices, facial recognition technology has found extensive applications in various domains such as mobile payments, identity authentication, and access control. To enhance the reliability of facial recognition systems, face anti-spoofing (FAS) systems have been introduced. However, the deployment of FAS models on the server-side, which often contain numerous parameters and integrate multiple models, has raised concerns regarding privacy and security. This is primarily due to the transmission of users' facial images over the network and their storage on the server, posing significant risks of privacy breaches. Moreover, the process of image transmission is time-consuming, resulting in a subpar user experience.

\begin{figure}[htb]
    \centering
    \includegraphics[width=0.5\textwidth]{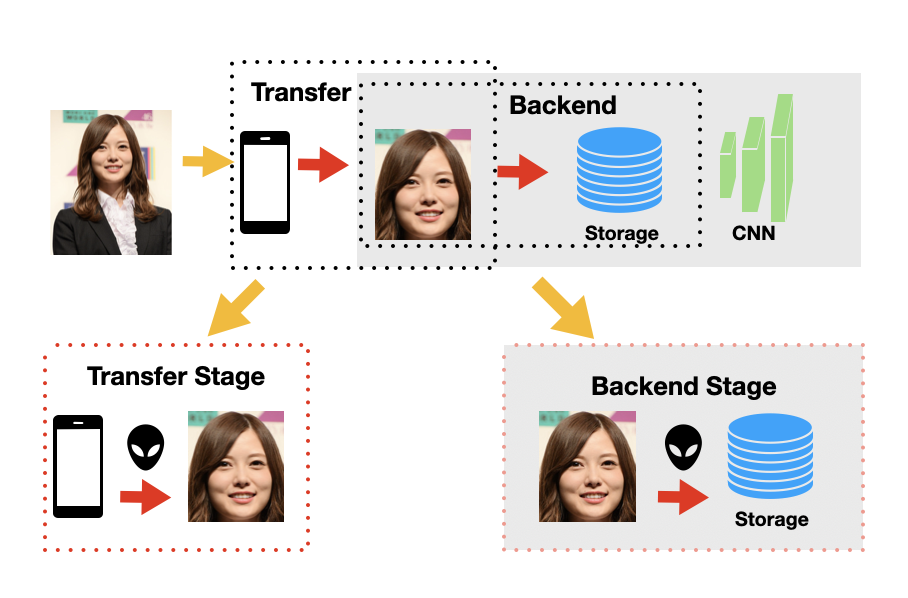}
    \caption{The traditional Face Anti-Spoofing system encompasses certain inherent vulnerabilities:1)Image Transmission Risk: One notable risk arises during the transmission phase, where facial images are susceptible to unauthorized interception and theft through illicit attacks.2)Storage Vulnerability: Another significant risk manifests during the storage phase, wherein facial images are potentially susceptible to unintended disclosure or leakage.}
    \label{intro}
\end{figure}

To address the aforementioned concerns, we propose a face anti-spoofing model that leverages pure facial skin patches. In contrast to the conventional approach of transmitting the entire face photo, which carries the risk of privacy breaches, our model focuses on decoupling texture features and structural features by selecting specific areas of facial skin that do not include identifiable facial features as input. These areas can be accurately located with the guidance of facial landmarks. Through extensive observation and testing, we have determined that the encryption and decryption of images constitute a significant portion of the overall algorithm's processing time. Since the extracted facial skin patches do not retain any identity-related information, the need for image encryption and decryption operations can be eliminated.


We utilized the CelebA-Spoof\cite{CelebA-Spoof} dataset for our research due to its comprehensive annotations and diverse range of scenes. Our initial preprocessing step involves cropping and aligning the face images. Facial patches are then extracted based on facial landmarks obtained using the Facemesh\cite{lugaresi2019mediapipe} algorithm. In order to enhance the model's generalization performance, we randomly select two patches from the set of all available patches and input them into two separate sub-branches. The features extracted by convolutional neural networks (CNN) are fused together and fed into a softmax layer to generate the final predictions.

To assess the effectiveness and robustness of our proposed scheme, we performed a series of experiments from various perspectives, including accuracy and latency. The outcomes demonstrate that our pure patch-based approach achieves high accuracy in detecting spoofing attacks while maintaining low time latency, particularly in terms of image transmission. To ensure a seamless user experience, we have developed an Android application on the client-side for extracting facial skin patches. On the server-side, we deployed the aforementioned network model. The entire process is completed in less than 100ms, allowing for real-time detection and ensuring a smooth user experience.

Our contributions can be summarized as follow:
\begin{itemize}
    \item We propose a Convolutional Neural Network (CNN) based approach for face anti-spoofing, focusing specifically on the utilization of facial skin patches for detecting spoofing attacks.
    \item As no sensitive information is transmitted or stored during the process, the encryption and decryption module can be safely eliminated, resulting in a significant speed improvement for the overall process.
    \item We have implemented our approach by designing a demonstration on an Android device. This demo showcases the face anti-spoofing function, which can accurately detect spoofing attacks with a latency of less than 100ms.
    
\end{itemize}

\begin{figure}[htb]
    \centering
    \includegraphics[width=0.5\textwidth]{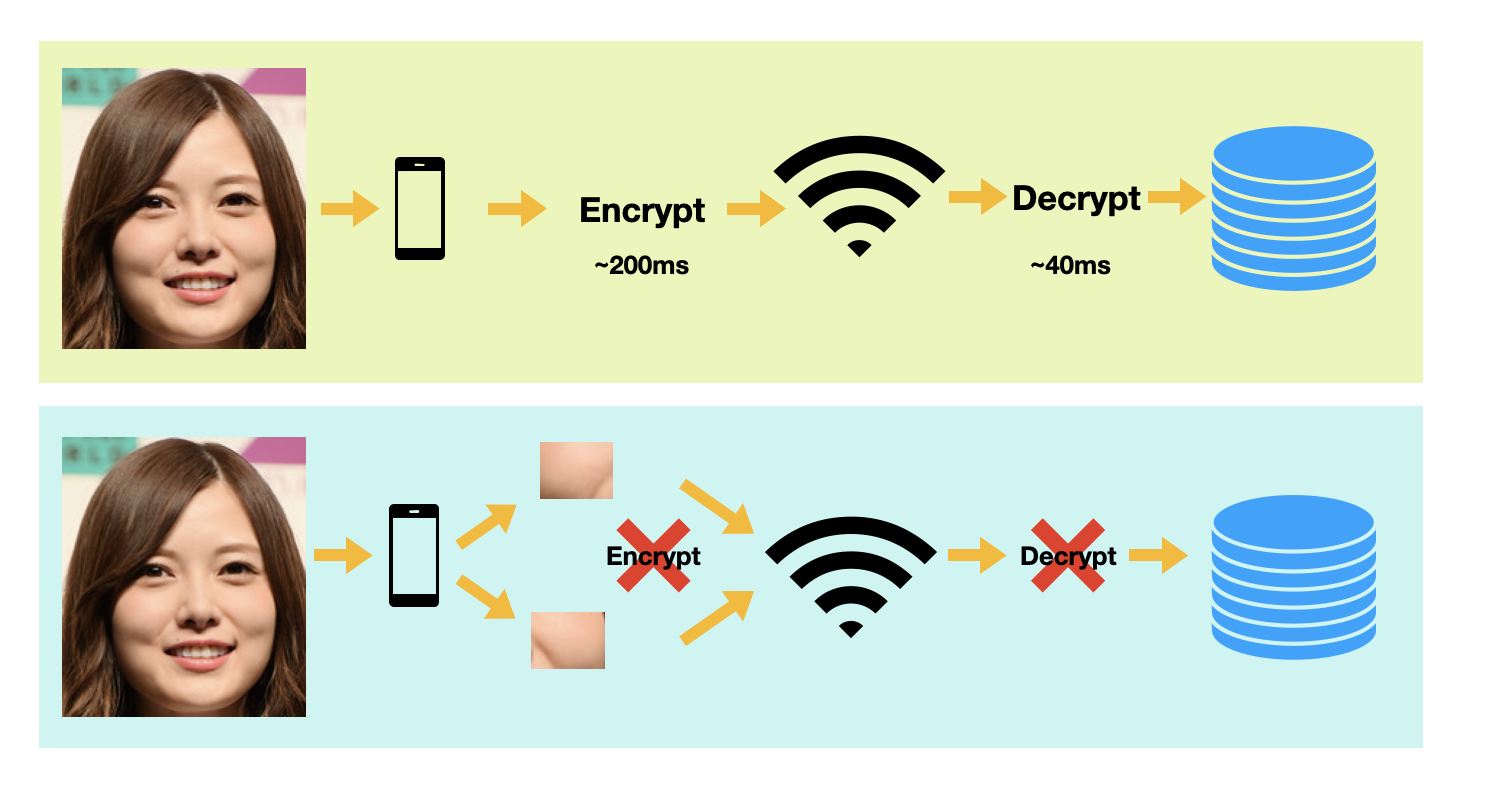}
    \caption{A comparative analysis of the previous workflow and our proposed approach reveals the following distinction: While the previous workflow necessitated the use of encryption and decryption processes to ensure privacy and security, our proposed approach eliminates the requirement for such operations.  }
    \label{encry}
\end{figure}

\section{Related work}

\subsection{Face Anti-Spoofing}
A large amount of feature-based methods have been proposed to tackle the face anti-spoofing problem. Handcrafted features, such as color, texture, can be leveraged as input to design a model. \cite{freitas2012lbp} propose an LBP-based face anti-spoofing approach, which improved the HTER from $15.16\%$ to $7.60\%$; \cite{komulainen2013context} proposed a context-based approach; SIFT\cite{komulainen2013context}, HoG\cite{patel2016secure} are also utilized as features to protect FRS.
\subsubsection*{\textbf{CNN based methods}}
Recently, deep learning has shown dominant performance in face antispoofing field. Several CNN-based approaches have outperformed the traditional methods. \cite{li2016original}extracts the deep partial features from the convolutional neural network (CNN) to distinguish the real faces from fake ones; Multi-scales has been leveraged to detect fake face images\cite{tulyakov2016self}; Auxiliary information, such as rPPG\cite{liu2018learning}, is leveraged to improve the performance of the face anti-spoofing system.

\begin{figure}[htb]
    \centering
    \includegraphics[width=0.5\textwidth]{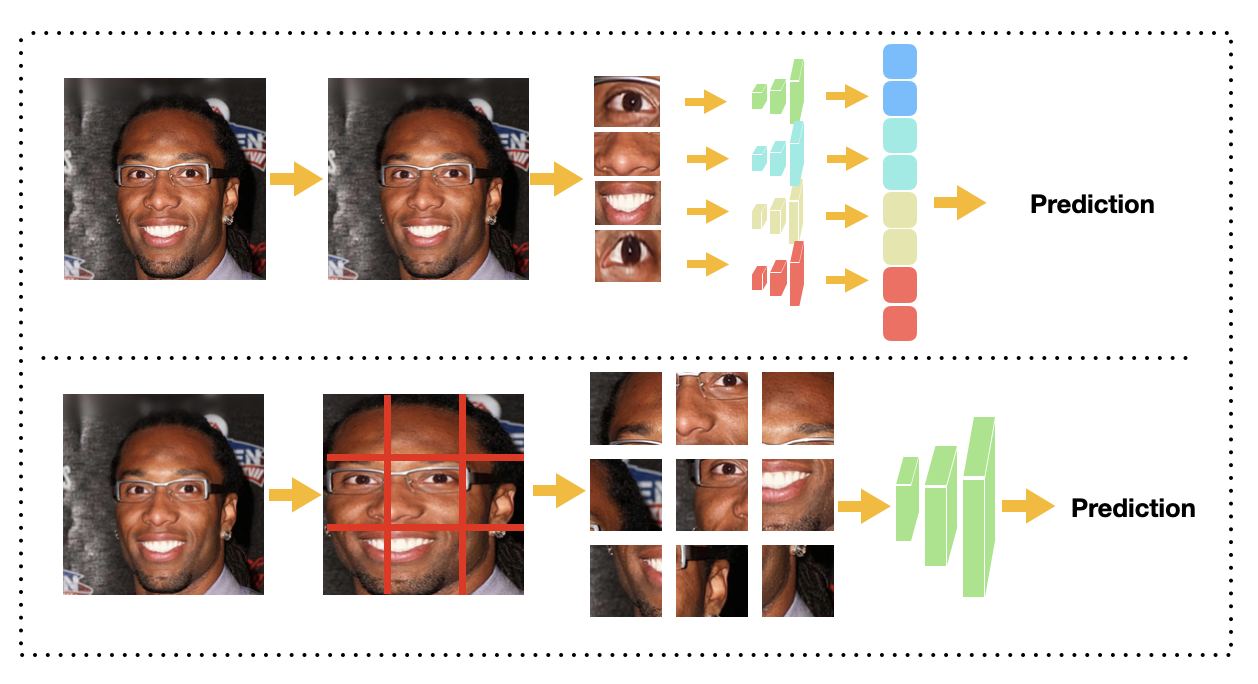}
    \caption{The utilization of conventional patch-based techniques for face anti-spoofing can be broadly categorized into two approaches: face feature-based (top row) and split-combine (bottom row).}
    \label{previous}
\end{figure}

\subsection{Patch-based Approach}

Patch-based  approach has been widely used in face anti-spoofing scenario. As shown in \textbf{Fig} \ref{previous}, \cite{visionlabs}propose a network which fuse face features extracted from face images. His team got first place in Chalearn 3D High-Fidelity Mask Face Presentation Attack Detection Challenge@ICCV2021. \cite{youtu} propose a patch-combinataion based approach, which splits face images into several sub-patches. These patches are than randomly combined forming a new image. \cite{youtu}assume that structure information and content information could be decoupled in this way. Both works have proved that patch-based approach are effective in face anti-spoofing scenario. However, these methods still have limitations. Besides, \cite{patchswap} proposed a patch-swap approach to boost FAS. Though above approaches extract patches from original face images, these images still contain rich identity information. In practical usage, these pieces of information carry potential risks of privacy breaches.

\begin{figure*}[htb]
    \centering
    \includegraphics[width=0.95\textwidth]{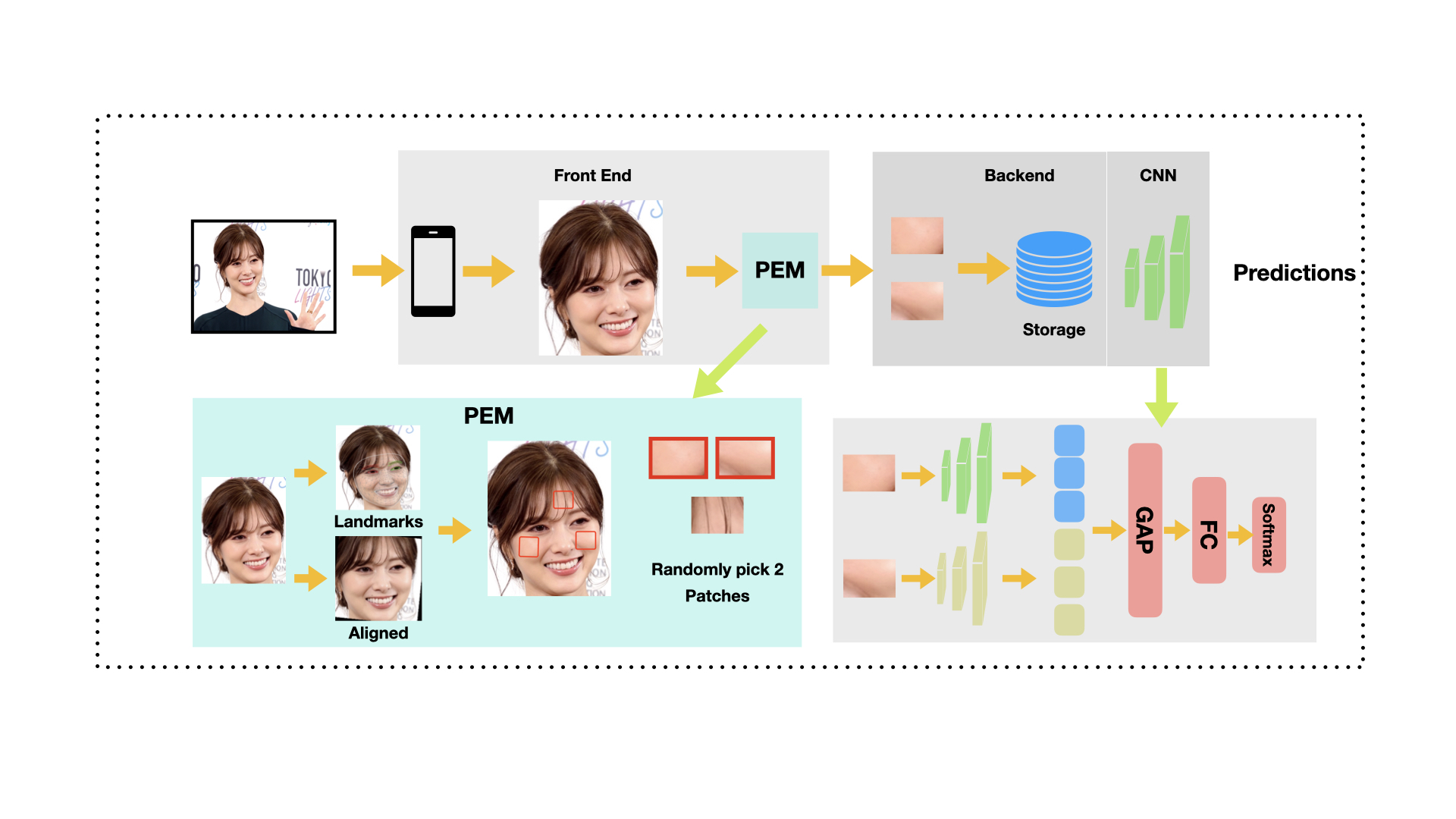}
    \caption{Our approach can be visualized in the following diagram, which comprises two main components: the front end and the back end. \textbf{Front End}: The input image undergoes a face detection process, resulting in the extraction of the face region; The cropped face is then fed into the PEM, which extracts skin patches from the face region. \textbf{Back End}:The CNN in the back end is responsible for processing the input patches and making predictions based on the learned features. }
    \label{intro}
\end{figure*}
\section{Motivation}

Through our observations, we have identified that the primary risks associated with live detection are primarily attributed to the image transmission and storage processes. As depicted in \textbf{Fig} \ref{encry}, the encryption and decryption operations consume approximately 240 ms within the entire process. In contrast, the main function, i.e., the Convolutional Neural Network (CNN) inference, only requires 20 ms. Nonetheless, it is crucial to employ encryption and decryption techniques to safeguard sensitive data that may contain private information. Additionally, the storage of facial images on the server side poses the risk of potential theft. To summarize, transmitting facial images can result in substantial delays and raise privacy concerns. The need for encryption and decryption steps can be circumvented if we succeed in decoupling the identity content from the biometric characteristics.


Previous studies have demonstrated the potential to separate the structure and texture components of face images in face anti-spoofing applications. In the context of patch-based face anti-spoofing algorithms, there are two main approaches. As depicted in \textbf{Fig} \ref{previous}, the first approach involves extracting patch images of specific facial features (such as eyes, nose, and mouth) as input features. The second approach entails dividing the face into multiple sub-patches. However, both approaches still carry the risk of privacy disclosure. Additionally, the first approach necessitates the use of four distinct models, which can impose a considerable computational burden.

To address these challenges, we conducted extensive research on attack scenarios and identified significant differences in the characteristics of attack materials used across various scenarios. Different attack materials exhibit distinctive image features, including reflection, texture, and others. Building on our previous work, we have demonstrated that pure facial patches can be effectively utilized as inputs for various deep learning classification tasks. In light of this, we hypothesize that facial skin patches can also be adapted for face anti-spoofing purposes.


\section{Method}

\subsection{Formulation}

\textbf{Face anti-spoofing} can be formulated as a classification task, where $1$ and $0$ annotate bona-fide and attack respectively.
Given an image $I$, our task is to design a model $\phi$ to make predictions as equation  \textbf{(1,2,4)}. A threshold $H$ can be set dynamically to judge the final label of the output.

\textbf{Traditional face anti-spoofing process} is as follow: portrait image $I_{p}$ is captured by mobile devices. Due to its sensitivity and privacy, images need to be encrypted during transmission and decrypted on the server side. The decrypted images are than fed into CNN model for final prediction. Total time consumption can be calculated as equation \textbf{(3)}.
\begin{equation}
    \hat{I}=\psi(I_{p})
\end{equation}
\begin{equation}
    I_{p} = \delta(\hat{I})
\end{equation}
\begin{equation}
    T_{total} = T_{trans}+T_{encry}+T_{decry}+T_{infer}
\end{equation}

\begin{equation}
    Prediction=\begin{cases}
    1, & \phi(I)>H\\ 
    0, & \phi(I)<H
    \end{cases}
\end{equation}
\begin{table*}[htb]
    \centering
    \resizebox{\textwidth}{20mm}{
    \begin{tabular}{cccccccccc}
    \toprule
         Method&\multicolumn{3}{c}{MSU} &\multicolumn{3}{c}{Youtu}&\multicolumn{3}{c}{Replay-attack\cite{replay-attack}} \\
         \cmidrule(lr){2-4} \cmidrule(lr){5-7} \cmidrule{8-10}\\
         &APCER(\%)&BPCER(\%)&ACER(\%)&APCER(\%)&BPCER(\%)&ACER(\%)&APCER(\%)&BPCER(\%)&ACER(\%)\\ 
         \midrule
         LBP\cite{chingovska2012effectiveness} &8.4&12.6&10.5&13.7&22.4&18.1&11.5&18.4&15.0\\
         color texture\cite{color-texture}&7.3&9.4&8.4&11.3&13.5&12.4&8.3&13.8&11.1\\
         FaceDs \cite{faceds}&4.2&5.3&4.5&9.3&7.5&8.4&7.8&9.7&8.5\\
         FASNet \cite{FASNet}&5.7&6.4&6.1&8.7&9.3&9.0&6.7&11.3&9.0\\
         Binary CNN \cite{binarycnn}&2.3&1.9&2.1&6.5&5.3&5.9&5.5&7.3&6.4\\
        CDC\cite{oulu}&0.8&1.6&\textbf{1.2}&2.4&1.2&\textbf{1.8}&2.1&1.7&\textbf{1.9}\\
         Ours &2.4&2.2&2.3&3.2&2.4&2.8&3.5&3.1&3.3\\
         \bottomrule
    \end{tabular}}
    \vspace{0.5cm}
    \caption{Experiments Results. Bold font indicate best performance.}
    \label{tab:result}
\end{table*}

\subsection{Data Acquisition}
Training a deep learning classification model requires a large amount of well-annotated data. However, obtaining human face images is costly and labor  intensive. 
CelebA-Spoofing is selected as the source images due to following advantages:
\begin{itemize}
    \item Large volume.
    \item Coverage of multiple attack types
    \item Free
    \item High quality
\end{itemize}
In each iteration of the process, the RetinaFace\cite{deng2020retinaface} is employed to crop the face region from the original image. Subsequently, the cropped face image is inputted into the facemesh model to acquire facial landmarks. These landmarks are then utilized to identify skin patches that do not overlap with any facial features. Specifically, we extract the aforementioned skin patches that do not contain any sensitive or personally identifiable information. Our data collection efforts encompassed the gathering of more than 10,000 patches, consisting of bona-fide samples as well as various types of attack instances.

\begin{figure}[htb]
    \centering
    \includegraphics[width=0.5\textwidth]{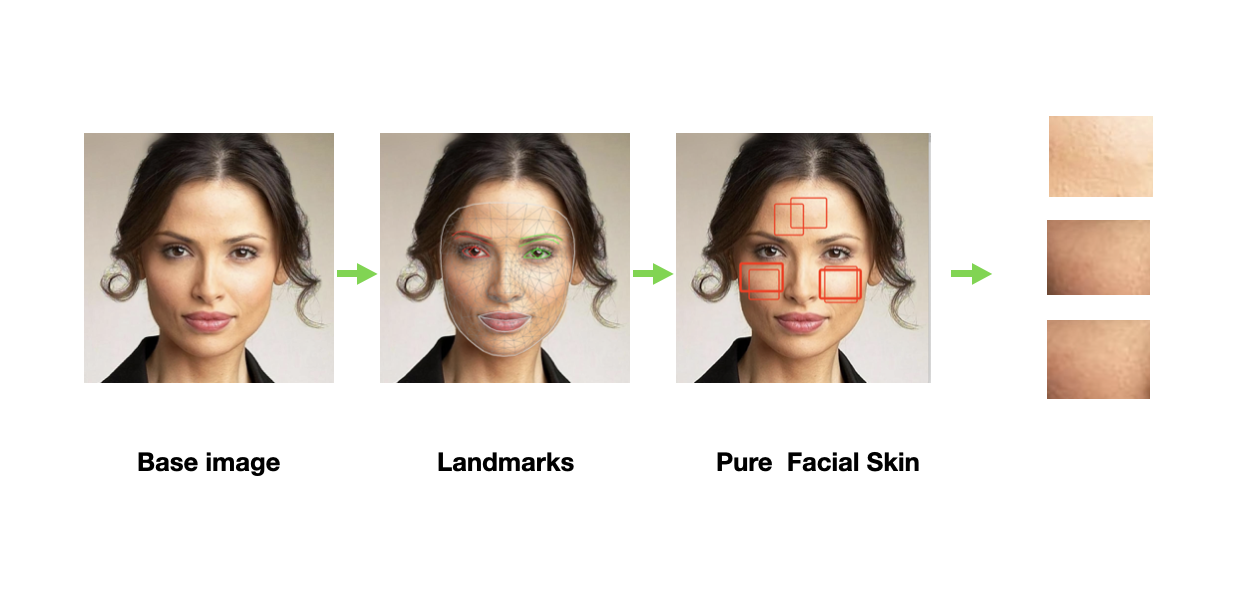}
    \caption{The utilization of conventional patch-based techniques for face anti-spoofing can be broadly categorized into two approaches: face feature-based (top row) and split-combine (bottom row).}
    \label{previous}
\end{figure}

\begin{table*}[htb]
    \centering
    \begin{tabular}{|c|c|c|c|c|c|}
    \hline
         \diagbox{approach}{duration}& transmission(ms)& encry(ms)& 
         decry(ms)&inference(ms)&total(ms)\\
         \hline
         ours& 53&0 &0&34&87\\
         \hline
         traditional&42&197&47&28&314\\
         \hline
    \end{tabular}
    \vspace{0.5cm}
    \caption{Duration comparison between traditional approach and our proposed approach.}
    \label{tab:Latency}
\end{table*}

\subsection{PEM}

In order to extract facial patches from face images, we present a scientific approach called the Patch-Extracted Module (PEM). The PEM utilizes facial landmark detection, guided by the facemesh algorithm \cite{lugaresi2019mediapipe}, to identify the landmarks on the input face images. Subsequently, the face images are aligned to ensure the extraction of high-quality facial patches.

For the selection of candidate regions, we specifically focus on the left cheek, right cheek, and chin areas, as they typically lack distinct facial features. To enhance the diversity of the extracted patches, we introduce a random shifting of the coordinates pertaining to the selected patches. From the various alternative patches obtained through this process, we choose two patches to further enrich the dataset.
\subsection{CNN}
To facilitate the fusion of features from two patches, we propose a novel double-branch architecture. The architectural design, as depicted in \textbf{Figure} \ref{intro}, comprises two branches that process each patch individually. These branches employ a convolutional neural network (CNN) backbone to extract features from their respective input patches. The obtained features are then concatenated to enhance the overall representational capacity of the network.

In contrast to employing a fully connected (FC) layer directly, we incorporate a global average pooling layer in order to reduce the total number of parameters. By adopting this approach, we can achieve effective dimensionality reduction without sacrificing the expressiveness of the network.
\section{Experiments}
We perform experiments to assess accuracy and latency, while also conducting ablation experiments to evaluate the influence of the number of branches. 
\subsection{Configurations}
Details of configurations are demonstrated as \textbf{table} \ref{tab:config}. 
\begin{table}[htb]
    \centering
    \begin{tabular}{|c|c|}
    \hline
     para    & value \\
    \hline
     GPU    &  Tesla T4\\
     \hline
     CPU    &  Intel Xeon 5218 \\
     \hline
     platform & Pytorch\\
     \hline
     loss function &  cross-entropy\\
     \hline
     optimizer & SGD\\
     \hline
     learning rate & 0.01\\
     \hline
     batch size & 64\\
     \hline
     epoch & 100\\
     \hline
     transform & Random Horizon filp, color jitter, \\
     \hline
    \end{tabular}
    \vspace{0.5cm}
    \caption{Configurations of experiments}
    \label{tab:config}
\end{table}

\subsection{Metric}

To evaluate the performance of models in face anti-spoofing, several metrics are commonly used, including the Attack Presentation Classification Error Rate (APCER), Bonafide Presentation Classification Error Rate (BPCER), and the Average Classification Error Rate (ACER). These metrics can be calculated as follows:
\begin{equation}
    APCER = FP/(TN+TP)
\end{equation}
\begin{equation}
    BPCER = FN/(FN+TP)
\end{equation}
\begin{equation}
    ACER = (APCER+BPCER)/2
\end{equation}

\begin{itemize}
    \item TP: The correct classification of a spoofed (fake) face as spoofed. 
    \item TN: the correct classification of a genuine face as genuine.
    \item FP: the model incorrectly classifies a genuine face as spoofed
    \item FN: the incorrect classification of a spoofed face as genuine.
\end{itemize}
\subsection{Results}
\subsubsection{Accuracy}
Here, we select Rose-Youtu\cite{youtu}, MSU\cite{msu} and Mobile-Replay\cite{replay-attack} as our test dataset.
 As depicted in \textbf{Table} \ref{tab:result}, 
 Our proposed model exhibits noteworthy performance in comparison to various algorithms when evaluated on three distinct datasets. Notably, the CDC algorithm emerges as the top performer across all three datasets. However, our model offers the advantage of being lightweight in comparison to CDC, making it highly convenient for deployment in the backend infrastructure. 

Furthermore, our model achieves results that are comparable to state-of-the-art approaches such as FaceDs and FASNet, while surpassing the performance of traditional algorithms like LBP and Color Texture by a significant margin. This outcome highlights the efficacy of our proposed model in addressing face anti-spoofing challenges. Additionally, our algorithm demonstrates stability and robustness across the aforementioned datasets, further affirming its reliability in different scenarios.
\subsubsection{Latency}
Total time consumption encompasses image transmission, image encryption and decryption, and model inference. We evaluated the latency of both the traditional approach(resnet34) and our proposed approach(two streams resnet34), and the results are as \textbf{table} \ref{tab:Latency}. Through The transmission and inference time consumption of our proposed approach is slightly greater than that of traditional one, total latency of our approach is only around 28\% of traditional one. The system benefit from discarding encryption and decryption components, which speeds up the whole process.  
\subsection{Ablation Study}
To assess the impact of the number of network branches on performance, we conducted ablation experiments. The experiments results demonstrate that with the number of network branches increases, the performance of the network gradually improves. To strike a balance between model size and accuracy, we finally pick 2-branch as our model structure. Besides, results from row 2 and row 4 indicate that random-picked patch can enhance the performance slightly.  
\begin{table}[htb]
    \centering
    \begin{tabular}{ccccc}
    \toprule
      Vanilla   & 2-branch & 3-branch & Rand patch& ACER(\%)\\
      \midrule
        \checkmark & & & & 5.78\\
        &\checkmark &  & &4.24\\ 
        & &\checkmark  &&\textbf{4.18}\\
        & &  & &4.24\\
        &\checkmark &  &\checkmark &4.26\\
    \bottomrule
    \end{tabular}
    \vspace{0.5cm}
    \caption{Ablation results. Bold indicates best performance.}
    \label{tab:my_label}
\end{table}


While the temporal duration of transmission and inference in our approach exhibits a minor increment compared to the conventional method, the aggregate latency is noticeably reduced owing to the absence of encryption and decryption procedures.

\section{Conclusion}
In this study, we present a novel mobile privacy-enhancing Face Anti-Spoofing framework that utilizes exclusive facial skin patches as input features, devoid of any encryption-decryption components. Our approach exhibits a significantly reduced time delay, approximately one-fourth, in comparison to conventional methods. 

\bibliographystyle{plain}

\bibliography{ref}
\end{document}